\documentclass{article} 

\usepackage{iclr2027_arxiv,times}

\usepackage{amsmath,amsfonts,bm}

\def\eqref#1{equation~\ref{#1}}

\def\1{\bm{1}}

\DeclareMathAlphabet{\mathsfit}{\encodingdefault}{\sfdefault}{m}{sl}
\SetMathAlphabet{\mathsfit}{bold}{\encodingdefault}{\sfdefault}{bx}{n}

\usepackage{helvet}
\usepackage{courier}
\usepackage{inconsolata}
\usepackage{wrapfig}
\usepackage{amsmath}
\usepackage{amssymb}
\usepackage{amsfonts}
\usepackage{mathrsfs}

\usepackage{booktabs}
\usepackage{multirow}

\usepackage{graphicx}
\usepackage{subcaption}

\usepackage[table]{xcolor}

\usepackage{algorithm}
\usepackage{algorithmic}

\usepackage{tikz}
\usetikzlibrary{
    shapes.geometric,
    arrows.meta,
    arrows,
    positioning
}

\usepackage{textcomp}
\usepackage{microtype}

\usepackage{natbib}

\usepackage{url}
\usepackage[
    colorlinks=true,
    citecolor=blue,
    linkcolor=blue,
    urlcolor=blue
]{hyperref}

\usepackage{breakurl}
\usepackage{pifont}
\def\BibTeX{{\rm B\kern-.05em{\sc i\kern-.025em b}\kern-.08em
T\kern-.1667em\lower.7ex\hbox{E}\kern-.125emX}}

\usepackage{graphicx}
\definecolor{rowgray}{gray}{0.94}

\title{GroupMask: Layer-Adaptive Group-wise Sparsity for Semi-Structured LLM Pruning\\
}

\author{
\textbf{Zhengao Li}$^{1,4,*}$ \quad
\textbf{Shuoqiu Li}$^{1,*}$ \quad
\textbf{Xiaofang Zhang}$^{2}$ \quad
\textbf{Yukai Jin}$^{1}$ \quad
\textbf{Gokcen Kestor}$^{3}$
\\
\textbf{Yanfu Zhang}$^{2}$ \quad
\textbf{Yiming Zeng}$^{4}$ \quad
\textbf{Bin Ren}$^{2}$ \quad
\textbf{Chuxu Zhang}$^{4}$ \quad
\textbf{Shangqian Gao}$^{1,\dagger}$
\\[0.6em]
$^{1}$Florida State University
\qquad
$^{2}$College of William \& Mary
\\
$^{3}$Barcelona Supercomputing Center
\qquad
$^{4}$University of Connecticut
\\[0.1em]
$^{*}$Equal contribution.
\qquad
$^{\dagger}$Corresponding author: \texttt{sg24bi@fsu.edu}
}
\begin{document}

\maketitle


\begin{abstract}
Semi-structured pruning compresses large language models (LLMs) while keeping a regular sparse structure, but the prevailing N:M pattern fixes the same local sparsity ratio in every layer.
Layer-adaptive sparsity allocation improves unstructured pruning, yet it has been reported to be less effective under N:M sparsity, leaving open whether adaptive allocation is of limited value for semi-structured pruning in general or only under the fine-grained N:M pattern.
We examine this question with group-level sparsity, which partitions each weight matrix into regular groups, retains or prunes each group as a whole, and allows each layer's sparsity ratio to vary under a global budget.
We propose \textbf{GroupMask}, which generates the group selectors of all layers with a lightweight hypernetwork, relaxes them with a Gumbel-Sigmoid parameterization and a straight-through estimator, and learns them through sparsity-budget regularization and self-distillation while keeping the pretrained weights frozen.
On LLaMA-2-7B at 50\% sparsity with the same $1\times256$ group size, learned layer-adaptive allocation reduces WikiText-2 perplexity from 10.02 to 8.30 and raises the average zero-shot accuracy from 0.455 to 0.496 relative to a uniform per-layer ratio.
GroupMask obtains the lowest WikiText-2 perplexity on LLaMA-2-7B and the highest average zero-shot accuracy with Alpaca calibration among the evaluated baselines on five LLaMA and Qwen models.
Our code is available at \url{https://github.com/ZhengaoLi/GroupMask}.
\end{abstract}

\section{Introduction}
Large Language Models (LLMs) have improved performance across a wide range of natural language processing tasks by scaling parameters, training data, and computation~\cite{opt,llama2,llmsurvey}.
This scale also incurs high computational and memory costs, which complicates deployment on resource-constrained hardware and in latency-sensitive applications.
Model compression aims to reduce these costs while preserving the general capabilities of LLMs.

Existing compression techniques for LLMs include pruning~\cite{mag2_ft1_unstr4,deepcomp_mag3_ft2_unstr5,sparsegpt_hand2,wanda_hand3}, knowledge distillation~\cite{distill_o}, quantization~\cite{frantar2022gptq}, and low-rank or modular compression~\cite{wang2025svd,lin2025modegpt}.
We focus on pruning, which removes redundant parameters directly from pretrained models.
According to the granularity of sparsity, pruning methods can be divided into unstructured, structured, and semi-structured pruning.
Unstructured pruning removes individual weights and offers the most flexibility, which helps preserve model quality~\cite{obd_unstr1_sec1,obs_unstr2_sec2,sparsegpt_hand2,wanda_hand3}; however, its irregular sparsity leads to inefficient memory access and limited acceleration on GPUs without specialized sparse kernels~\cite{spinfer_gpu2}.
Structured pruning removes larger units such as channels, hidden dimensions, attention heads, or layers, producing models that run efficiently on standard hardware~\cite{networkslimming_str2,llmpruner_str3_hidden1,shortgpt_ft5_str5_layer1_hand4,slicegpt_str6_hidden2_hand5,disp_str8_hidden3_learn6}, but its coarse granularity typically causes a larger loss in quality, especially when a large fraction of parameters is removed.

Semi-structured pruning lies between these two extremes~\cite{twotwofour_semi1,channelperm_semi2,learnnm_semi3}.
Existing semi-structured methods mainly adopt N:M sparsity, which retains N non-zero weights in every M consecutive weights; the 2:4 case is natively accelerated on NVIDIA Ampere and later GPUs~\cite{twotwofour_semi1,learnnm_semi3}.
Learnable methods such as MaskLLM and ProxSparse further optimize N:M masks end-to-end instead of relying on hand-crafted importance scores~\cite{maskllm_semi4_learn9_gumbel2,liu2025proxsparse}.
Regardless of how the mask is obtained, N:M sparsity fixes the local sparsity ratio to $1-N/M$ in every layer and module.

For unstructured pruning, allocating sparsity non-uniformly across layers has been shown to improve over a uniform ratio, with the per-layer ratios either derived from outlier statistics~\cite{yin2024owl} or learned~\cite{xu2024besa}.
Under N:M sparsity, however, the same benefit is not observed: \citet{liu2025proxsparse} report that non-uniform layer-wise ratios become less effective under fine-grained N:M patterns, since critical weights may still be removed within each block.
Two explanations are consistent with this observation: layer-adaptive allocation may be of limited value for semi-structured LLM pruning in general, or the fine-grained local constraint of N:M may prevent it from taking effect.

\begin{figure}[t]
    \vspace{-12pt}
    \centering
    \includegraphics[
        width=\linewidth,
        trim=2 2 2 2,
        clip
    ]{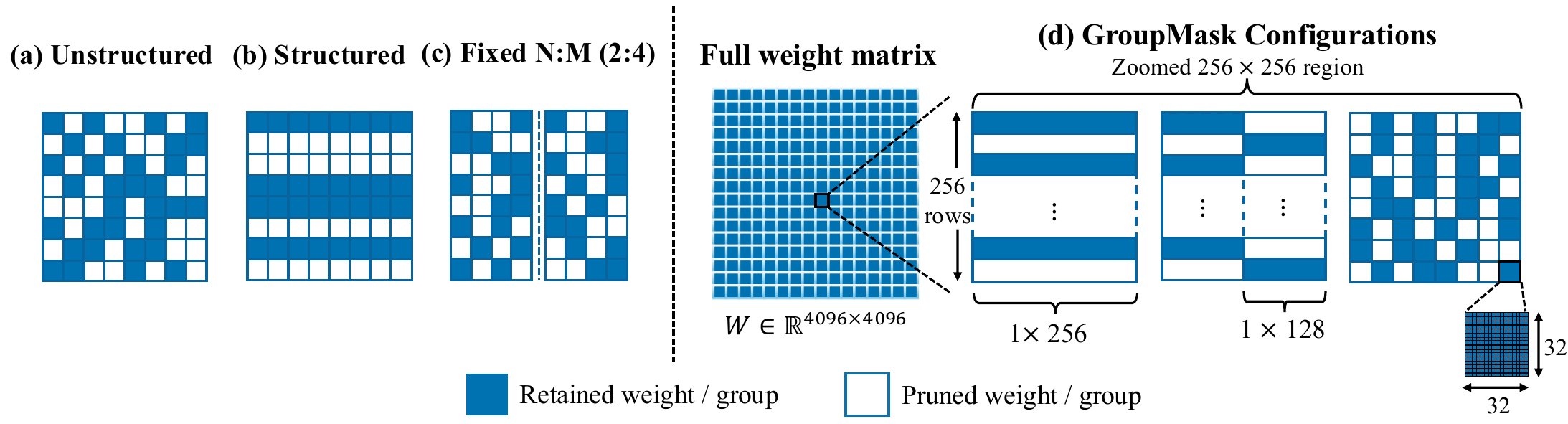}
    \caption{
        Comparison of pruning granularities:
        (a) unstructured pruning,
        (b) structured pruning,
        (c) fixed N:M sparsity, and
        (d) group-level sparsity with different group shapes.
    }
    \label{fig:pruning_compare}
    \vspace{-20pt}
\end{figure}

Distinguishing these explanations requires a sparse pattern that is regular but imposes no fixed local ratio.
We therefore study group-level sparsity, illustrated in Figure~\ref{fig:pruning_compare}(d).
Whereas N:M sparsity keeps exactly N weights in every M-wide window, group-level sparsity partitions each weight matrix into regular groups, such as $1\times256$, $1\times128$, or $32\times32$, and retains or prunes each group as a whole.
The sparsity ratio of a layer is then determined by how many of its groups are retained, and can fall above or below the global budget.

Selecting groups under a global budget is a combinatorial problem, since the decision for each group depends on its own importance, on the layer it belongs to, and on the budget consumed by the other layers.
We address this problem with \textbf{GroupMask}.
A lightweight hypernetwork produces selector logits for the groups of all layers, which are converted into binary selectors through a Gumbel-Sigmoid parameterization with a straight-through estimator (STE).
A regularizer on the global kept-parameter ratio enforces the budget, and self-distillation from the dense model provides the training signal~\cite{distill_o,learn4_distill1}.
The pretrained weights remain frozen, so the compressed model is a sparse sub-network of the original LLM.

Under the same $1\times256$ group structure on LLaMA-2-7B at 50\% sparsity, learned layer-adaptive allocation reduces WikiText-2 perplexity from 10.02 to 8.30 and raises the average zero-shot accuracy from 0.455 to 0.496 relative to a uniform per-layer ratio (Section~\ref{sec:adaptive_vs_uniform}).
Adaptive allocation is therefore beneficial once the fixed local ratio is removed, which suggests that the limitation observed under N:M sparsity is tied to the pattern rather than to adaptive allocation itself.
In perplexity, GroupMask outperforms SparseGPT~\cite{sparsegpt_hand2}, Wanda~\cite{wanda_hand3}, LLM-Pruner~\cite{llmpruner_str3_hidden1}, SliceGPT~\cite{slicegpt_str6_hidden2_hand5}, MoDeGPT~\cite{lin2025modegpt}, and DISP-LLM~\cite{disp_str8_hidden3_learn6} on LLaMA-2-7B; in zero-shot accuracy, with Alpaca calibration, it obtains the highest average among the evaluated baselines on all five LLaMA and Qwen models.

Our contributions are summarized as follows:
\begin{itemize}
    \item \textbf{Layer-adaptive allocation beyond N:M.}
    To the best of our knowledge, we are the first to learn group-level semi-structured masks for billion-parameter LLMs with frozen pretrained weights.
    With this pattern, we show that learned layer-adaptive allocation outperforms a uniform per-layer ratio at the same group granularity, in contrast to prior observations under N:M sparsity.

    \item \textbf{GroupMask.}
    We propose a hypernetwork-based method that generates the group selectors of all layers jointly and learns them with sparsity-budget regularization and self-distillation, without updating the pretrained weights.

    \item \textbf{Empirical evaluation and analysis.}
    We evaluate GroupMask on five LLaMA and Qwen models with perplexity and six zero-shot benchmarks, and analyze the learned allocation, which retains 36.6\% of attention parameters and 56.6\% of MLP parameters in LLaMA-2-7B and keeps only about 8\% of the query and key groups in layers 22--31.
\end{itemize}

\section{Methodology}
\label{sec:methodology}

\subsection{Notation and Group Sparsity Pattern}

Consider a transformer-based language model, and let $W^l \in \mathbb{R}^{d^l_{\text{out}} \times d^l_{\text{in}}}$, $l=1,\dots,L$, denote its $L$ target linear layers, where $d^l_{\text{out}}$ and $d^l_{\text{in}}$ are the output and input dimensions. To impose a regular sparse structure, we partition each weight matrix into non-overlapping groups of size $g^l_{\text{out}} \times g^l_{\text{in}}$, giving $G^l_{\text{out}} = d^l_{\text{out}} / g^l_{\text{out}}$ and $G^l_{\text{in}} = d^l_{\text{in}} / g^l_{\text{in}}$ groups along the output and input dimensions. Each group is retained or pruned as a whole according to a group-level selector $B^l \in \{0,1\}^{G^l_{\text{out}} \times G^l_{\text{in}}}$, where $B^l_{uv}=1$ retains group $(u,v)$ and $B^l_{uv}=0$ prunes it. We refer to each $W^l$ as a linear layer, so layer-adaptive allocation assigns a separate sparsity ratio to each projection (e.g., the query or down projection of a Transformer block) and can vary across both depth and projection type.

The full-resolution binary mask and the corresponding pruned weight matrix are
given by
\begin{equation}
    M^l = B^l \otimes
    \mathbf{1}_{g^l_{\mathrm{out}} \times g^l_{\mathrm{in}}},
    \qquad
    \hat{W}^l = W^l \odot M^l ,
    \label{eq:mask_apply}
\end{equation}
where $\otimes$ denotes the Kronecker product. The layer-wise and overall
sparsity ratios are defined as
\begin{equation}
s^l =
1-
\frac{
\sum_{u=1}^{G^l_{\mathrm{out}}}
\sum_{v=1}^{G^l_{\mathrm{in}}} B^l_{uv}
}{
G^l_{\mathrm{out}}G^l_{\mathrm{in}}
},
\qquad
s =
\frac{
\sum_l s^l d^l_{\mathrm{out}}d^l_{\mathrm{in}}
}{
\sum_l d^l_{\mathrm{out}}d^l_{\mathrm{in}}
}.
\label{eq:sparsity}
\end{equation}

Compared with element-wise masks, group-level selectors reduce the number of binary decision variables by a factor of $g^l_{\mathrm{out}} g^l_{\mathrm{in}}$, e.g., 256 for the default $1\times256$ groups.

\subsection{Hypernetwork-based Mask Generation}

Rather than scoring groups with a fixed heuristic such as weight magnitude, GroupMask generates the selectors with a hypernetwork $h_{\theta}$. Its input $z=(z_1,\dots,z_L)$ is a fixed random sequence with one vector per target layer. A bidirectional GRU processes this sequence, and a layer-specific output head maps the GRU output of layer $l$ to $G^l_{\mathrm{out}}G^l_{\mathrm{in}}$ selector logits, which are binarized into
\begin{equation}
B^l = h^l_{\theta}({z}),
\qquad
B^l \in \{0,1\}^{G^l_{\mathrm{out}} \times G^l_{\mathrm{in}}}.
\end{equation}

Because $B^l$ is discrete, it cannot be optimized directly by gradient descent. We use a Gumbel-Sigmoid relaxation~\cite{gumbel} with a straight-through estimator (STE)~\cite{bengio2013estimating}: the forward pass uses binary selectors, and the backward pass propagates gradients through the continuous relaxation to $\theta$. At evaluation time, the Gumbel noise is removed and the selectors are obtained by thresholding. The selectors are expanded and applied to the pretrained weights according to Eq.~(\ref{eq:mask_apply}). Since the selectors of all layers are produced by one network and trained under a single global budget, the per-layer sparsity ratios are determined jointly rather than fixed in advance.


\subsection{Sparsity Budget Regularization}

To enforce a target sparsity $s_t$, we regularize the global sparsity ratio $s$ defined in Eq.~(\ref{eq:sparsity}) with a log-ratio objective
$\mathcal{L}_{\mathrm{reg}}=\lambda\left|\log(s/s_t)\right|$, where $\lambda$ controls the regularization strength. Unlike distance-based penalties such as the squared error, this objective depends only on the ratio $s/s_t$ and assigns the same penalty to $s=c\,s_t$ and $s=s_t/c$. It is computed on the discrete selectors $B^l$, so it reflects the exact sparsity of the applied masks. Because $\mathcal{L}_{\mathrm{reg}}$ constrains only the global ratio $s$, the layer-wise ratios $s^l$ may deviate from $s_t$, which is what allows layer-adaptive allocation. The uniform variant in Section~\ref{sec:adaptive_vs_uniform} instead applies the same regularizer to each $s^l$ separately.

\begin{figure}[t]
    \vspace{-34pt}
    \centering
    \includegraphics[width=0.6\linewidth]{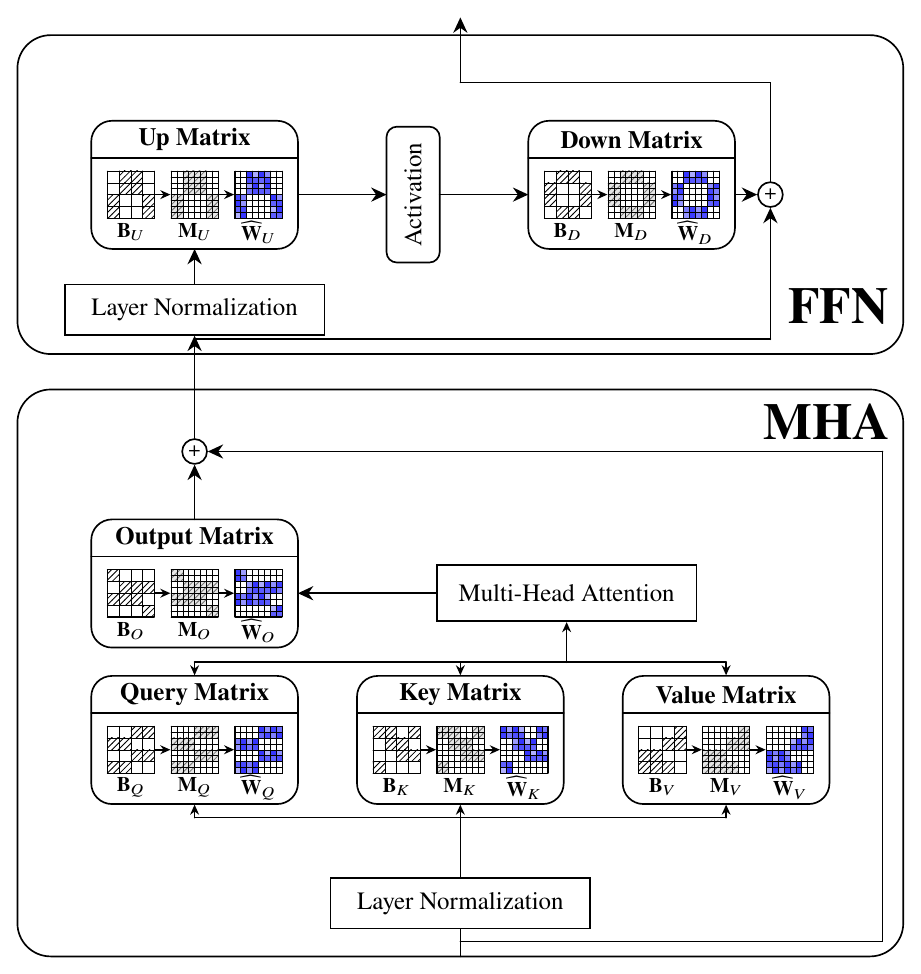}
    \caption{Structural configuration of our group-sparse pruning framework within a single Transformer block. For each linear projection, the mask generation module yields a structural group selector matrix $B$, which is expanded into a full-resolution binary mask $M$ to determine the pruned weight $\hat{W}^l$ via element-wise multiplication.}
    \label{fig:block_structure}
    \vspace{-16pt}
\end{figure}

\subsection{Self-Distillation Optimization Framework}

Fig.~\ref{fig:block_structure} illustrates the overall optimization framework.
The hypernetwork first generates group-level selector matrices $B^l$ for the
target projections, which are expanded into full-resolution masks and applied
to the pretrained weights according to Eq.~(\ref{eq:mask_apply}). The resulting
sparse model is optimized through self-distillation while the pretrained
weights remain frozen.

Unlike conventional knowledge distillation with separate teacher and student
models, our framework uses a single pretrained model with dynamically switched
sparsity masks. The teacher forward pass uses the dense model by setting all
selector matrices to $B^l=\mathbf{1}$, whereas the student forward pass applies
the generated selectors $B^l$ to obtain the sparse model. Since the pretrained
weights remain frozen throughout training, gradients are propagated only to
the hypernetwork parameters $\theta$.

To preserve the predictive behavior of the dense model, we minimize the
cross-entropy between the teacher and student output distributions:
\begin{equation}
    \mathcal{L}_{\mathrm{KD}}
    =
    -\frac{1}{N}
    \sum_{t=1}^{N}
    \sum_{v \in \mathcal{V}}
    p_T^{(t,v)}
    \log p_S^{(t,v)},
\end{equation}
where $N$ denotes the number of valid tokens and $\mathcal{V}$ denotes the
vocabulary. This objective is equivalent to minimizing the forward KL
divergence from the teacher distribution to the student distribution up to an
additive constant:
\begin{equation*}
\begin{aligned}
    \mathrm{KL}(p_T \,\|\, p_S)
    &=
    \underbrace{
    - \sum_v p_T(v)\log p_S(v)
    }_{\text{cross-entropy}}
    -
    \underbrace{
    \left(
    - \sum_v p_T(v)\log p_T(v)
    \right)
    }_{H(p_T)} .
\end{aligned}
\end{equation*}
The entropy term $H(p_T)$ depends only on the teacher distribution and is
constant with respect to the student. Therefore, minimizing
$\mathcal{L}_{\mathrm{KD}}$ is equivalent to minimizing the forward KL
divergence without explicitly computing this constant term.

The final training objective combines self-distillation and sparsity
regularization:
\begin{equation}
    \mathcal{L}_{\mathrm{total}}
    =
    \mathcal{L}_{\mathrm{KD}}
    +
    \mathcal{L}_{\mathrm{reg}}.
    \label{eq:loss}
\end{equation}
Together, these objectives guide GroupMask to preserve the predictive behavior
of the dense model while satisfying the target sparsity constraint.


\section{Experiments}

We conduct experiments to answer five research questions for GroupMask: 
(1) Can GroupMask preserve language modeling quality after pruning? 
(2) Does GroupMask generalize across different LLM families and downstream tasks? 
(3) Does layer-adaptive allocation outperform a uniform per-layer ratio under the same group pattern? 
(4) How does group granularity affect pruning quality? 
(5) What is the contribution of the hypernetwork-based mask generator?

\subsection{Settings}

\noindent \textbf{Models.}
We evaluate LLaMA-2-7B, LLaMA-2-13B, LLaMA-3-8B, Qwen3-8B, and Qwen3-14B, which cover two model families and 7B to 14B parameters.

\noindent \textbf{Implementation.}
We implement our method with PyTorch~\cite{paszke2019pytorchimperativestylehighperformance} and Hugging Face Transformers~\cite{wolf-etal-2020-transformers}.
The pretrained weights are frozen, and no post-pruning fine-tuning is applied.
Unless stated otherwise, we use $1\times256$ groups, 50\% sparsity, and WikiText-2 calibration.
For zero-shot evaluation, we additionally learn masks with Alpaca calibration; the two settings are denoted GroupMask-Wiki and GroupMask-Alpaca.
Further details are given in Appendix~\ref{app:implementation}.

\noindent \textbf{Datasets.}
We use WikiText~\cite{merity2016pointer} and Alpaca~\cite{alpaca} as calibration data.
We measure language modeling quality by perplexity on WikiText-2, and zero-shot performance on six benchmarks~\cite{eval-harness}: ARC-Challenge~\cite{allenai:arc}, ARC-Easy~\cite{allenai:arc}, BoolQ~\cite{clark2019boolq}, HellaSwag~\cite{zellers2019hellaswag}, PIQA~\cite{bisk2019piqareasoningphysicalcommonsense}, and WinoGrande~\cite{sakaguchi2019winograndeadversarialwinogradschema}.
The average zero-shot score is the mean over the six tasks.

\noindent \textbf{Baselines.}
We compare with the one-shot 2:4 methods SparseGPT~\cite{sparsegpt_hand2} and Wanda~\cite{wanda_hand3}, and with the structured methods LLM-Pruner~\cite{llmpruner_str3_hidden1}, SliceGPT~\cite{slicegpt_str6_hidden2_hand5}, MoDeGPT~\cite{lin2025modegpt}, and DISP-LLM~\cite{disp_str8_hidden3_learn6}, all at 50\% sparsity.
We do not include MaskLLM~\cite{maskllm_semi4_learn9_gumbel2}: its reported LLaMA-2-7B result uses 512k training samples and 1,280 A100 GPU-hours, whereas GroupMask uses about 3.7 GPU-hours (Table~\ref{tab:app_hyperparameters}), so the two methods operate at compute budgets that are not comparable.

\subsection{Language Modeling Performance}

\noindent 
\begin{minipage}[t]{0.5\textwidth}
    \vspace{0pt} 
    Table~\ref{tab:ppl} reports WikiText-2 perplexity at 50\% sparsity.
    On LLaMA-2-7B, GroupMask with $1\times256$ groups obtains the lowest perplexity among the compared methods, 8.30 compared with 9.84 for DISP-LLM, a 15.7\% relative reduction.
    On LLaMA-2-13B, it obtains 7.19, second to DISP-LLM (7.11).
    Structured methods yield smaller dense matrices that run without sparse kernels, so this comparison is made at equal parameter sparsity rather than at equal inference cost.
    The $1\times256$ configuration outperforms $32\times32$ on both models (8.30 vs.\ 9.37 and 7.19 vs.\ 7.79).
    A $1\times256$ group contains 256 weights, whereas a $32\times32$ group contains 1,024, so the former makes four times as many selection decisions per matrix; Section~\ref{sec:ablation} examines group granularity further.
\end{minipage}%
\hfill 
\begin{minipage}[t]{0.44\textwidth} 
    \vspace{0pt}
    \centering
    \footnotesize 
    \setlength{\tabcolsep}{3pt} 
    \renewcommand{\arraystretch}{1.1} 
    
    \captionof{table}{WikiText-2 perplexity ($\downarrow$) of LLaMA-2 models at 50\% sparsity.}
    \captionsetup{justification=centering}
    \label{tab:ppl}
    \vspace{4pt}
    
    \hspace*{-1.5em}
    \begin{tabular}{@{}lcccc@{}} 
    \toprule
    Method & Sparsity type & W Update & 7B & 13B \\
    \midrule
    Dense & -- & -- & 5.12 & 4.57 \\
    \midrule
    LLM-Pruner & Structured & \ding{55} & 109.43 & 49.71 \\
    SliceGPT   & Structured & \ding{51} & 24.82 & 20.57 \\
    MoDeGPT    & Structured & \ding{51} & 11.88 & 8.95 \\
    DISP-LLM   & Structured & \ding{55} & 9.84 & \textbf{7.11} \\
    \midrule
    SparseGPT & 2:4 & \ding{51} & 10.17 & 8.32 \\
    Wanda     & 2:4 & \ding{55} & 11.02 & 8.27 \\
    \midrule
    \shortstack[l]{GroupMask \\ ($1\times256$)} & Group & \ding{55} & \textbf{8.30} & \underline{7.19} \\
    \shortstack[l]{GroupMask \\ ($32\times32$)} & Group & \ding{55} & \underline{9.37} & 7.79 \\
    \bottomrule
    \end{tabular}
    \vspace{4pt}
\end{minipage}
\vspace{0pt} 

\subsection{Zero-shot Downstream Performance}

\begin{table*}[h]
\vspace{-12pt}
\centering
\small
\setlength{\tabcolsep}{4pt}
\renewcommand{\arraystretch}{1.05}
\caption{
Zero-shot downstream performance under 50\% sparsity ($\uparrow$).
}
\vspace{-8pt}
\label{tab:llama2_downstream_main}
\resizebox{0.95\textwidth}{!}{%
\begin{tabular}{l|l|cccccc|c}
\toprule
\multirow{2}{*}{Model} & \multirow{2}{*}{Method}
& ARC-C
& ARC-E
& BoolQ
& Hella
& PIQA
& Wino
& \multirow{2}{*}{\textbf{Avg}}
\\
\cline{3-8}
&
& acc-norm
& acc-norm
& acc
& acc-norm
& acc-norm
& acc
&
\\
\midrule

\multirow{7}{*}{LLaMA-2-7B}
& Dense        & 0.462 & 0.745 & 0.778 & 0.760 & 0.788 & 0.692 & 0.704 \\
& LLM-Pruner$^{*}$~\cite{llmpruner_str3_hidden1}   & 0.239 & 0.304 & 0.454 & 0.276 & 0.525 & 0.499 & 0.383 \\
& MoDeGPT~\cite{lin2025modegpt}      & 0.293 & 0.375 & 0.621 & 0.458 & 0.610 & 0.604 & 0.493 \\
& DISP-LLM~\cite{disp_str8_hidden3_learn6}     & 0.260 & 0.393 & 0.629 & 0.433 & 0.607 & 0.529 & 0.475 \\
& SliceGPT~\cite{slicegpt_str6_hidden2_hand5}     & 0.237 & 0.348 & 0.378 & 0.325 & 0.536 & 0.531 & 0.393 \\
& \cellcolor{gray!8}GroupMask-Wiki        & \cellcolor{gray!8}0.261 & \cellcolor{gray!8}0.462 & \cellcolor{gray!8}0.588 & \cellcolor{gray!8}0.472 & \cellcolor{gray!8}0.644 & \cellcolor{gray!8}0.550 & \cellcolor{gray!8}0.496 \\
& \cellcolor{gray!20}GroupMask-Alpaca      & \cellcolor{gray!20}0.306 & \cellcolor{gray!20}0.542 & \cellcolor{gray!20}0.547 & \cellcolor{gray!20}0.496 & \cellcolor{gray!20}0.670 & \cellcolor{gray!20}0.577 & \cellcolor{gray!20}\textbf{0.523} \\
\midrule

\multirow{7}{*}{LLaMA-2-13B}
& Dense        & 0.492 & 0.775 & 0.805 & 0.794 & 0.805 & 0.721 & 0.732 \\
& LLM-Pruner$^{*}$~\cite{llmpruner_str3_hidden1}   & 0.240 & 0.343 & 0.481 & 0.309 & 0.577 & 0.493 & 0.407 \\
& MoDeGPT~\cite{lin2025modegpt}      & 0.325 & 0.487 & 0.649 & 0.537 & 0.630 & 0.680 & 0.551 \\
& DISP-LLM~\cite{disp_str8_hidden3_learn6}     & 0.323 & 0.502 & 0.639 & 0.530 & 0.641 & 0.572 & 0.534 \\
& SliceGPT~\cite{slicegpt_str6_hidden2_hand5}     & 0.249 & 0.370 & 0.378 & 0.344 & 0.555 & 0.555 & 0.409 \\
& \cellcolor{gray!8}GroupMask-Wiki        & \cellcolor{gray!8}0.312 & \cellcolor{gray!8}0.523 & \cellcolor{gray!8}0.623 & \cellcolor{gray!8}0.558 & \cellcolor{gray!8}0.676 & \cellcolor{gray!8}0.587 & \cellcolor{gray!8}0.546 \\

& \cellcolor{gray!20}GroupMask-Alpaca      & \cellcolor{gray!20}0.355 & \cellcolor{gray!20}0.620 & \cellcolor{gray!20}0.631 & \cellcolor{gray!20}0.571 & \cellcolor{gray!20}0.707 & \cellcolor{gray!20}0.578 & \cellcolor{gray!20}\textbf{0.577} \\
\midrule

\multirow{6}{*}{LLaMA-3-8B}
& Dense        & 0.533 & 0.777 & 0.814 & 0.792 & 0.808 & 0.729 & 0.742 \\
& MoDeGPT~\cite{lin2025modegpt}      & 0.315 & 0.453 & 0.669 & 0.410 & 0.628 & 0.509 & 0.497 \\
& DISP-LLM~\cite{disp_str8_hidden3_learn6}     & 0.279 & 0.432 & 0.556 & 0.468 & 0.639 & 0.553 & 0.488 \\
& SliceGPT~\cite{slicegpt_str6_hidden2_hand5}     & 0.211 & 0.303 & 0.378 & 0.297 & 0.517 & 0.503 & 0.368 \\
& \cellcolor{gray!8}GroupMask-Wiki        & \cellcolor{gray!8}0.261 & \cellcolor{gray!8}0.439 & \cellcolor{gray!8}0.622 & \cellcolor{gray!8}0.430 & \cellcolor{gray!8}0.635 & \cellcolor{gray!8}0.559 & \cellcolor{gray!8}0.491 \\
& \cellcolor{gray!20}GroupMask-Alpaca      & \cellcolor{gray!20}0.297 & \cellcolor{gray!20}0.524 & \cellcolor{gray!20}0.618 & \cellcolor{gray!20}0.453 & \cellcolor{gray!20}0.663 & \cellcolor{gray!20}0.545 & \cellcolor{gray!20}\textbf{0.517} \\
\midrule

\multirow{6}{*}{Qwen3-8B}
& Dense        & 0.564 & 0.809 & 0.866 & 0.749 & 0.777 & 0.676 & 0.740 \\
& MoDeGPT~\cite{lin2025modegpt}      & 0.263 & 0.428 & 0.594 & 0.350 & 0.607 & 0.504 & 0.457 \\
& DISP-LLM~\cite{disp_str8_hidden3_learn6}     & 0.316 & 0.500 & 0.624 & 0.459 & 0.631 & 0.559 & 0.515 \\
& SliceGPT~\cite{slicegpt_str6_hidden2_hand5}     & 0.229 & 0.327 & 0.378 & 0.331 & 0.545 & 0.534 & 0.391 \\
& \cellcolor{gray!8}GroupMask-Wiki        & \cellcolor{gray!8}0.290 & \cellcolor{gray!8}0.498 & \cellcolor{gray!8}0.628 & \cellcolor{gray!8}0.442 & \cellcolor{gray!8}0.630 & \cellcolor{gray!8}0.571 & \cellcolor{gray!8}0.510 \\
& \cellcolor{gray!20}GroupMask-Alpaca      & \cellcolor{gray!20}0.359 & \cellcolor{gray!20}0.612 & \cellcolor{gray!20}0.657 & \cellcolor{gray!20}0.469 & \cellcolor{gray!20}0.671 & \cellcolor{gray!20}0.563 & \cellcolor{gray!20}\textbf{0.555} \\
\midrule

\multirow{6}{*}{Qwen3-14B}
& Dense        & 0.602 & 0.828 & 0.894 & 0.788 & 0.798 & 0.729 & 0.773 \\
& MoDeGPT~\cite{lin2025modegpt}      & 0.275 & 0.447 & 0.636 & 0.368 & 0.620 & 0.532 & 0.480 \\
& DISP-LLM~\cite{disp_str8_hidden3_learn6}     & 0.365 & 0.570 & 0.453 & 0.551 & 0.669 & 0.616 & 0.537 \\
& SliceGPT~\cite{slicegpt_str6_hidden2_hand5}     & 0.235 & 0.349 & 0.378 & 0.355 & 0.553 & 0.576 & 0.408 \\
& \cellcolor{gray!8}GroupMask-Wiki        & \cellcolor{gray!8}0.312 & \cellcolor{gray!8}0.547 & \cellcolor{gray!8}0.640 & \cellcolor{gray!8}0.500 & \cellcolor{gray!8}0.647 & \cellcolor{gray!8}0.587 & \cellcolor{gray!8}0.539 \\
& \cellcolor{gray!20}GroupMask-Alpaca      & \cellcolor{gray!20}0.386 & \cellcolor{gray!20}0.628 & \cellcolor{gray!20}0.708 & \cellcolor{gray!20}0.521 & \cellcolor{gray!20}0.698 & \cellcolor{gray!20}0.592 & \cellcolor{gray!20}\textbf{0.589} \\
\bottomrule
\end{tabular}
}
\vspace{-12pt}
\parbox{0.98\textwidth}{
\footnotesize
$^{*}$LLM-Pruner is reported only on LLaMA-2 models because its available
implementation does not directly support the LLaMA-3 and Qwen3 architectures
evaluated here.
}
\end{table*}

Table~\ref{tab:llama2_downstream_main} reports zero-shot accuracy at 50\% sparsity; all GroupMask results use $1\times256$ groups.
With Alpaca calibration, GroupMask obtains the highest average accuracy among the evaluated baselines on all five models.
On LLaMA-2-7B, LLaMA-2-13B, and LLaMA-3-8B, its averages of 0.523, 0.577, and 0.517 exceed the strongest baseline, MoDeGPT, by 3.0, 2.6, and 2.0 percentage points.
On Qwen3-8B and Qwen3-14B, its averages of 0.555 and 0.589 exceed the strongest baseline, DISP-LLM, by 4.0 and 5.2 points.
With WikiText-2 calibration, GroupMask obtains the highest average on LLaMA-2-7B and Qwen3-14B and is within 0.6 points of the strongest baseline on the other three models.
Alpaca calibration yields a higher average than WikiText-2 calibration on all five models, with the largest difference on Qwen3-14B (0.539 to 0.589).

\subsection{Layer-Adaptive versus Uniform Allocation}
\label{sec:adaptive_vs_uniform}

\begin{table}[t]
\vspace{6pt}
\centering
\caption{Layer-adaptive versus uniform sparsity allocation on LLaMA-2-7B at 50\% sparsity. Both settings use the same $1\times256$ group structure, WikiText-2 calibration, and training schedule; the uniform setting constrains every linear layer to $s^l = 50\%$.}
\label{tab:adaptive_vs_uniform}
\begin{tabular}{lcccccccc}
\toprule
Allocation & PPL $\downarrow$ & ARC-C & ARC-E & BoolQ & Hella & PIQA & Wino & Avg $\uparrow$ \\
\midrule
Uniform  & 10.02 & 0.238 & 0.376 & \textbf{0.622} & 0.378 & 0.596 & 0.522 & 0.455 \\
Adaptive & \textbf{8.30} & \textbf{0.261} & \textbf{0.462} & 0.588 & \textbf{0.472} & \textbf{0.644} & \textbf{0.550} & \textbf{0.496} \\
\bottomrule
\end{tabular}
\vspace{-6pt}
\end{table}

To isolate the effect of layer-adaptive allocation, we compare GroupMask with a uniform variant that constrains every target linear layer to $s^l = 50\%$, while keeping the group structure, calibration data, and training schedule unchanged.
As shown in Table~\ref{tab:adaptive_vs_uniform}, adaptive allocation reduces WikiText-2 perplexity from 10.02 to 8.30 and raises the average zero-shot accuracy from 0.455 to 0.496, improving five of the six tasks; BoolQ decreases from 0.622 to 0.588.
Since the two settings differ only in whether the per-layer ratio is fixed or learned, the gap can be attributed to layer-adaptive allocation under the group-level pattern.
This result contrasts with the observation of \citet{liu2025proxsparse} that non-uniform layer-wise ratios become less effective under N:M sparsity.
The non-uniform ratios evaluated there differ from ours in two respects: they are derived from heuristics such as OWL~\cite{yin2024owl}, and OWL assigns one ratio per Transformer block, whereas GroupMask learns a ratio for each projection.
The comparison therefore suggests, but does not by itself establish, that the fine-grained N:M constraint is what limits the benefit of non-uniform ratios.
Section~\ref{sec:ablation} shows how the learned allocation departs from a uniform ratio: about half of its variation is across projection types, which block-level ratios cannot express, and deep attention projections receive much smaller budgets than MLP projections.

\begin{figure}[thb]

    \centering

    \begin{subfigure}[t]{0.32\linewidth}
        \centering
        \includegraphics[width=\linewidth]
        {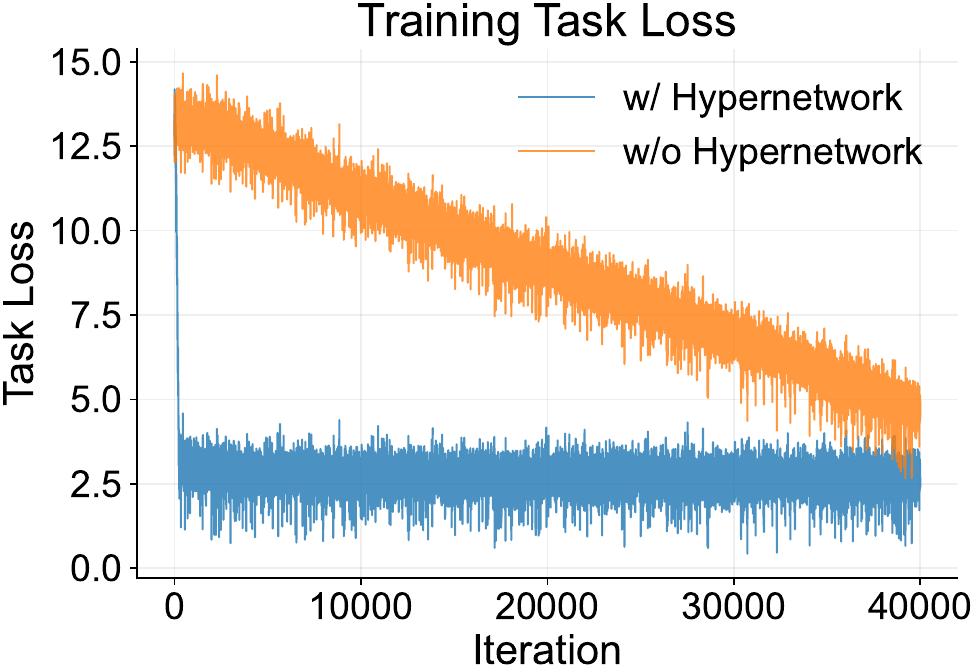}
        \caption{Hypernetwork optimization}
        \label{fig:task_loss_comparison}
    \end{subfigure}
    \hfill
    \begin{subfigure}[t]{0.32\linewidth}
        \centering
        \includegraphics[width=\linewidth]
        {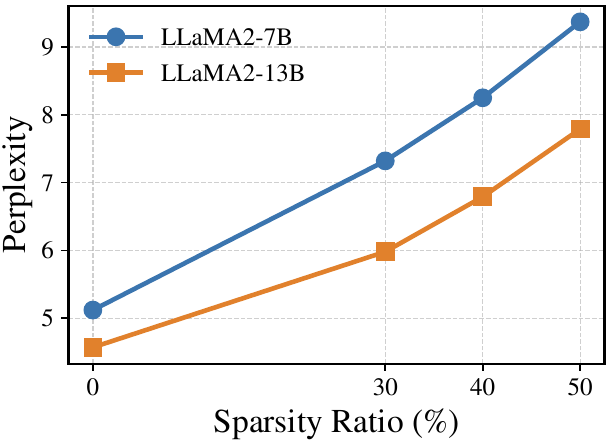}
        \caption{Perplexity}
        \label{fig:ppl_sensitivity}
    \end{subfigure}
    \hfill
    \begin{subfigure}[t]{0.32\linewidth}
        \centering
        \includegraphics[width=\linewidth]
        {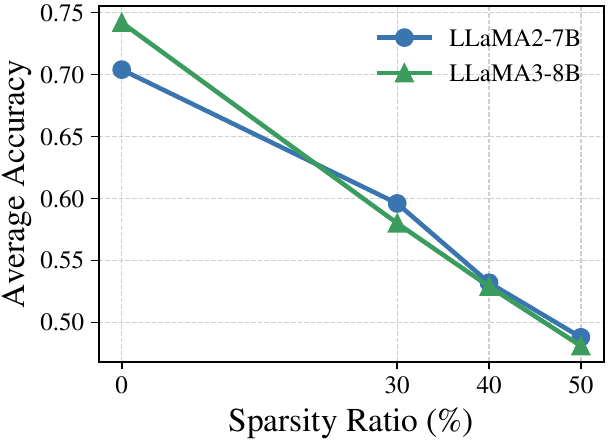}
        \caption{Zero-shot accuracy}
        \label{fig:sparsity_sensitivity}
    \end{subfigure}

    \caption{
    Ablation analysis of GroupMask.
    (a) Effect of the hypernetwork on mask optimization.
    (b--c) Perplexity and zero-shot performance under increasing sparsity.
    }
    \label{fig:ablation_curves}
    \vspace{-16pt}
\end{figure}

\subsection{Ablation Study}
\label{sec:ablation}

\begin{table}[t] 
\centering
\footnotesize 
\setlength{\tabcolsep}{4pt} 
\renewcommand{\arraystretch}{1.0} 

\caption{Zero-shot ablation on LLaMA-2-7B at 50\% sparsity ($\uparrow$).}
\label{tab:ablation}

\begin{tabular}{@{}llccccccc@{}}
\toprule
Setting & Cal. & ARC-C & ARC-E & BoolQ & Hella. & PIQA & Wino. & Avg. \\
\midrule

Dense & -- &
0.462 & 0.745 & 0.778 & 0.760 & 0.788 & 0.692 & 0.704 \\

\midrule
\multicolumn{9}{@{}l}{\textit{Hypernetwork ablation}} \\
\addlinespace[2pt] 

w/o HN & Wiki &
0.240 & 0.322 & 0.478 & 0.314 & 0.552 & 0.511 & 0.403 \\
      \rowcolor{rowgray} & Alp. &
0.242 & 0.283 & 0.528 & 0.269 & 0.522 & 0.479 & 0.387 \\

\midrule
\multicolumn{9}{@{}l}{\textit{Group granularity and calibration}} \\
\addlinespace[2pt]

$1\times128$ & Wiki &
0.269 & 0.441 & 0.631 & 0.469 & 0.642 & 0.560 & 0.502 \\
             \rowcolor{rowgray} & Alp. &
0.292 & 0.551 & 0.626 & 0.486 & 0.680 & 0.537 & \textbf{0.528} \\
\addlinespace[2pt]

$1\times256$ & Wiki &
0.261 & 0.462 & 0.588 & 0.472 & 0.644 & 0.550 & 0.496 \\
             \rowcolor{rowgray} & Alp. &
0.306 & 0.542 & 0.547 & 0.496 & 0.670 & 0.577 & \underline{0.523} \\
\addlinespace[2pt]

$16\times16$ & Wiki &
0.269 & 0.446 & 0.604 & 0.471 & 0.656 & 0.549 & 0.499 \\
             \rowcolor{rowgray} & Alp. &
0.288 & 0.524 & 0.628 & 0.459 & 0.669 & 0.542 & 0.518 \\
\addlinespace[2pt]

$32\times32$ & Wiki &
0.270 & 0.440 & 0.573 & 0.464 & 0.653 & 0.526 & 0.488 \\
             \rowcolor{rowgray} & Alp. &
0.279 & 0.481 & 0.624 & 0.465 & 0.656 & 0.541 & 0.508 \\
\addlinespace[2pt]

$64\times64$ & Wiki &
0.278 & 0.460 & 0.610 & 0.437 & 0.630 & 0.535 & 0.492 \\
             \rowcolor{rowgray} & Alp. &
0.271 & 0.476 & 0.602 & 0.453 & 0.655 & 0.541 & 0.499 \\
\addlinespace[2pt]

$128\times128$ & Wiki &
0.232 & 0.405 & 0.554 & 0.437 & 0.599 & 0.493 & 0.453 \\
               \rowcolor{rowgray} & Alp. &
0.247 & 0.435 & 0.624 & 0.433 & 0.620 & 0.541 & 0.483 \\

\bottomrule
\end{tabular}
\end{table}

We examine four design choices: (1) the hypernetwork, (2) group granularity, (3) calibration data, and (4) the sparsity ratio. We then analyze the allocation learned by GroupMask.

\noindent\textbf{Effect of the hypernetwork.}
We compare mask learning with and without the hypernetwork on LLaMA-2-7B.
In the setting without a hypernetwork, the selector logits are learnable parameters initialized from a standard normal distribution and passed through the same Gumbel-Sigmoid relaxation and STE, with the same regularizer and distillation objective.
With the hypernetwork, the task loss is lower throughout training (Figure~\ref{fig:task_loss_comparison}), and removing it reduces the average zero-shot accuracy from 0.496 to 0.403 with WikiText-2 calibration and from 0.523 to 0.387 with Alpaca calibration (Table~\ref{tab:ablation}).
The corresponding sparsity regularization loss is shown in Appendix~\ref{app:additional_speedup}.

\noindent\textbf{Group granularity and calibration data.}
As shown in Table~\ref{tab:ablation}, average accuracy tends to decrease as each group contains more weights.
With WikiText-2 calibration, it decreases from 0.502 for $1\times128$ to 0.453 for $128\times128$, although the trend is not strictly monotonic ($32\times32$: 0.488; $64\times64$: 0.492).
With Alpaca calibration, $1\times128$ obtains the highest average (0.528), followed by the default $1\times256$ (0.523).
For every group shape in Table~\ref{tab:ablation}, Alpaca calibration yields a higher average than WikiText-2 calibration, e.g., 0.496 to 0.523 for $1\times256$.

\noindent\textbf{Effect of sparsity ratio.}
Figures~\ref{fig:ppl_sensitivity} and~\ref{fig:sparsity_sensitivity} report perplexity and zero-shot accuracy at 30\%, 40\%, and 50\% sparsity with $32\times32$ groups and WikiText-2 calibration.
From 30\% to 50\% sparsity, the perplexity of LLaMA-2-7B increases from 7.32 to 9.37 and its average accuracy decreases from 0.596 to 0.488; the average accuracy of LLaMA-3-8B decreases from 0.580 to 0.481 (Appendix~\ref{app:sparsity}).

\begin{figure}[thb]
    \centering
    \includegraphics[width=\linewidth]{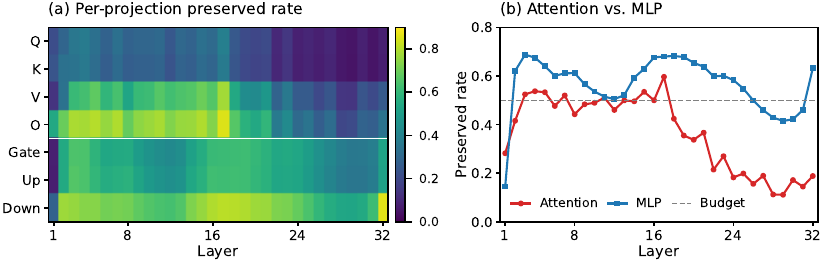}
    \vspace{-15pt}
    \caption{
    Learned preserved rates of LLaMA-2-7B at 50\% sparsity.
    (a) Preserved rate of each projection across layers.
    (b) Average preserved rates of attention and MLP projections;
    the dashed line marks the global 50\% budget.
    }
    \label{fig:layerwise_preserved_ratio}
    \vspace{0pt}
\end{figure}

\noindent\textbf{Kernel-level efficiency.}
Group-level sparsity has no native hardware support, so we implement a preliminary group-sparse SpMM kernel with CUDA/CUTLASS on an NVIDIA H100 using Hopper's Tensor Memory Accelerator (TMA), and compare it with dense cuBLAS in FP32 on a LLaMA-2 gate projection ($M=11008$, $K=4096$, $N=65536$).
With group size 64, the kernel reaches $1.09\times$ speedup at 50\% sparsity and $1.88\times$ at 90\% (Figure~\ref{fig:app_efficiency}); at 50\% sparsity, changing the group size has little effect on speedup (Figure~\ref{fig:app_speedup_group}).
These are kernel-level measurements; end-to-end acceleration of group-sparse LLMs requires further system support, which we leave to future work.

\noindent\textbf{Analysis of learned pruning structures.}
Figure~\ref{fig:layerwise_preserved_ratio} shows the preserved rate of each projection in the exported LLaMA-2-7B model.
The global budget is met (49.98\% of parameters preserved), but the allocation is non-uniform: attention projections retain 36.6\% of their parameters and MLP projections 56.6\%, and the down projection accounts for 30.5\% of all preserved parameters.
Three regularities appear.
First, functionally coupled projections receive nearly identical budgets: across layers, the preserved rates of Q and K, and those of Gate and Up, have a Pearson correlation of 0.994 and differ by less than 0.05.
Second, within attention, V retains about 2.2 times as many parameters as Q or K, and O retains the most, so the learned masks prune the query--key pathway more heavily than the value--output pathway.
Third, the allocation varies with depth: in layers 3--17, attention and MLP receive comparable budgets, whereas in layers 22--31, Q and K keep about 8\% of their groups while the down projection keeps more than 53\%.
This depth-dependent reduction of attention is consistent with reports that many deep attention layers in LLMs can be removed with limited quality loss~\citep{he2024matterstransformersattentionneeded,shortgpt_ft5_str5_layer1_hand4}. In GroupMask, the pattern is learned from the distillation objective rather than specified by a pruning rule, and it is expressed per projection rather than per Transformer block.
Across the 224 projections in Figure~\ref{fig:layerwise_preserved_ratio}, projection type accounts for about half of the variance in preserved rates and depth for about one third, and roughly 40\% of all target parameters lie in projections pruned beyond 50\% sparsity, which neither 2:4 sparsity nor PATCH~\cite{patch_semi5_learn10} can represent.

\section{Related Work}

\noindent \textbf{Pruning for LLMs.}
Unstructured pruning removes individual weights and preserves model quality well~\cite{obd_unstr1_sec1,obs_unstr2_sec2,sparsegpt_hand2,wanda_hand3}, but its irregular sparsity limits practical acceleration~\cite{spinfer_gpu2}.
Structured pruning removes channels, hidden dimensions, attention heads, or layers~\cite{llmpruner_str3_hidden1,slicegpt_str6_hidden2_hand5,shortgpt_ft5_str5_layer1_hand4,disp_str8_hidden3_learn6}, yielding dense models that run efficiently on standard hardware at the cost of coarser pruning decisions.
Post-training methods such as SparseGPT and Wanda show that LLMs can be pruned without full retraining~\cite{sparsegpt_hand2,wanda_hand3}, while distillation-guided pruning replaces recovery fine-tuning with supervision from the dense model~\cite{distill_o,learn4_distill1}.

\noindent \textbf{Semi-structured sparsity.}
N:M sparsity retains N non-zero weights in every M consecutive weights, and the 2:4 case is natively accelerated on NVIDIA Ampere and later GPUs~\cite{twotwofour_semi1,channelperm_semi2,learnnm_semi3}.
Learnable N:M methods optimize the mask end-to-end, either through Gumbel-softmax sampling over candidate patterns~\cite{maskllm_semi4_learn9_gumbel2} or through proximal regularization~\cite{liu2025proxsparse}.
PATCH~\cite{patch_semi5_learn10} partitions weights into tiles and makes each tile either dense or 2:4 sparse, which lets the sparsity ratio vary across layers; since each tile is at most 50\% sparse, however, no layer can exceed 50\% sparsity, and at a 50\% global budget the pattern reduces to uniform 2:4 sparsity.
Block pruning has been studied for BERT-scale encoders, where block masks are learned by movement pruning together with weight updates during task-specific fine-tuning~\cite{lagunas2021block}, and GPU kernels for block-sparse weights were developed earlier~\cite{gray2017blocksparse}.
GroupMask instead learns group-level masks for billion-parameter decoder-only LLMs in a task-agnostic, post-training setting with frozen weights, and generates the masks of all layers jointly, so that each layer's sparsity ratio can fall above or below the global budget.

\noindent \textbf{Layer-wise sparsity allocation.}
For unstructured pruning, non-uniform layer-wise sparsity improves over a uniform ratio, whether the ratios are derived from outlier statistics~\cite{yin2024owl} or learned~\cite{xu2024besa}.OWL assigns one ratio per Transformer block and reports that assigning ratios to individual layers degrades performance.
Under N:M sparsity, \citet{liu2025proxsparse} observe that non-uniform layer-wise ratios become less effective, because critical weights may still be removed within each block.
We examine this discrepancy by comparing adaptive and uniform allocation under the same group-level pattern.

\noindent \textbf{Learnable masks.}
Learnable pruning methods parameterize pruning decisions and optimize them with gradient-based relaxations such as movement scores~\cite{movement_learn3}, Gumbel-based sampling~\cite{learn5_gumbel1}, or decaying masks~\cite{learn7_semi7}.
DISP-LLM~\cite{disp_str8_hidden3_learn6} uses a hypernetwork to generate structural pruning decisions for LLMs.
GroupMask follows this line by generating the group-level selectors of all layers with a shared hypernetwork, trained through self-distillation while the pretrained weights stay frozen.

\section{Conclusion}
We revisited layer-adaptive sparsity allocation for semi-structured LLM pruning, which prior work found to be less effective under the N:M pattern.
With group-level sparsity, which keeps a regular structure but leaves each layer's sparsity ratio free, learned adaptive allocation reduces WikiText-2 perplexity from 10.02 to 8.30 relative to a uniform ratio under the same group structure on LLaMA-2-7B, suggesting that the earlier observation is tied to the N:M pattern.
GroupMask learns this allocation with a hypernetwork-generated group mask, sparsity-budget regularization, and self-distillation, without updating the pretrained weights, and with Alpaca calibration it obtains the highest average zero-shot accuracy among the evaluated baselines on five LLaMA and Qwen models.
The learned masks remove most query and key groups in deep layers while retaining more of the MLP projections.
\noindent\textbf{Limitations.}
First, the comparison with N:M sparsity is indirect: the non-uniform ratios evaluated under N:M are heuristic and assigned per Transformer block, whereas ours are learned per projection; a heuristic allocation under group-level sparsity would further separate the two factors.
Second, the adaptive-versus-uniform comparison is conducted on LLaMA-2-7B only.
Third, the highest zero-shot averages are obtained with Alpaca calibration; with WikiText-2 calibration, GroupMask is best on two of the five models.
Finally, group-level sparsity currently lacks native hardware support, our kernel measurements are preliminary, and our experiments cover sparsity of at most 50\% and models of up to 14B parameters; end-to-end acceleration and co-design with sparse kernels remain future work.
\clearpage
\subsection*{AI use statement}

Generative AI tools were used during this work to assist with research
methodology and experimental design discussions, software development and
debugging, analysis and interpretation of experimental results, and
preparation of the manuscript. They were also used for drafting and editing
text, improving clarity and organization, formatting LaTeX tables and
equations, and providing feedback on figures and presentation.
All AI-assisted research decisions, implementations, experimental results,
and manuscript content were reviewed and verified by the authors. Experimental
results reported in this paper were obtained from our implementations and
experiments rather than generated by AI tools. The authors take full
responsibility for the methodology, results, claims, and final content of the
paper.
\subsection*{Reproducibility statement}

We provide implementation and experimental details to support reproducibility.
The methodology section specifies the group sparsity formulation, mask
generation, sparsity regularization, and self-distillation objective, while
the appendix reports training hyperparameters, calibration settings, additional
results, mask export procedures, and sparse-kernel evaluation details.
Our implementation is available in the anonymous repository linked in the
abstract.
\subsection*{Ethics statement}

This work studies model compression for large language models and does not
involve human subjects or the collection of personal data. Our experiments
use publicly available pretrained models and benchmark datasets.

\bibliographystyle{plainnat}
\bibliography{custom}

\clearpage
\appendix

\section{Implementation Details}
\label{app:implementation}

This section provides additional implementation details for reproducing our experiments. 
All experiments are implemented with PyTorch and Hugging Face Transformers. 
For each pruned model, we keep the original pretrained LLM weights frozen and optimize only the hypernetwork-based mask generator. 
No LoRA adaptation, recovery fine-tuning, or post-pruning weight update is applied.

\noindent \textbf{Training objective.}
We use the same objective as described in Sec.~\ref{sec:methodology}, and provide additional implementation details here.
During mask learning, the dense pretrained model is used as the teacher, and the masked model is used as the student.
The teacher logits are computed with pruning masks disabled and are detached from the computation graph.
The student logits are computed with group-sparse masks enabled.
The pretrained weights remain frozen throughout this process, and only the hypernetwork parameters used to generate group-level masks are updated.

\noindent \textbf{Calibration data.}
We use WikiText and Alpaca as calibration datasets for learning pruning structures. 
The two calibration settings are denoted as GroupMask-Wiki and GroupMask-Alpaca, respectively. 
Unless otherwise specified, the calibration dataset seed is fixed to 42. 
Input sequences are tokenized by the corresponding model tokenizer, and the maximum calibration sequence length is set to 2048.

\noindent \textbf{Optimization hyperparameters.}
Table~\ref{tab:app_hyperparameters} summarizes the major hyperparameters used in our experiments. 
For experiments with different sparsity ratios, we only change the target sparsity parameter while keeping the other hyperparameters unchanged. 
For group-shape ablations, we only change the group dimensions and keep the optimization setting fixed.

\begin{table}[h]
\centering
\small
\caption{Main hyperparameters used for hypernetwork-based mask learning.}
\label{tab:app_hyperparameters}
\begin{tabular}{lc}
\toprule
Hyperparameter & Value \\
\midrule
Optimizer & AdamW / AdamW-8bit \\
Learning rate & $1\times10^{-3}$ \\
Weight decay & 0.05 \\
Adam $\beta$ & $(0.9, 0.999)$ \\
Batch size & 1 \\
Sequence length & 2048 \\
Training steps & 40,000 \\
Training time (LLaMA-2-7B) & $\sim$3.7 GPU-hours \\
Training hardware & 1 $\times$ NVIDIA DGX B200 \\
Checkpoint interval & 10,000 \\
Calibration seed & 42 \\
Temperature $T$ & 0.4 \\
Regularization coefficient $\lambda$ & 16 \\
Default sparsity ratio & 50\% \\
Default group shape & $1\times256$ \\
Precision & bfloat16 \\
FSDP & Enabled when needed \\
Post-pruning finetuning & No \\
LoRA adaptation & No \\
\bottomrule
\end{tabular}
\vspace{-10pt}
\end{table}

\noindent \textbf{Group shapes and sparsity ratios.}
The main experiments use the default group configuration unless otherwise specified. 
For group-granularity analysis, we additionally evaluate $16\times16$, $1\times128$, $1\times256$, $32\times32$, $64\times64$, and $128\times128$ group structures. 
For sparsity sensitivity analysis, we evaluate 30\%, 40\%, and 50\% sparsity. 
The target sparsity is controlled by the mask regularization objective, while the original model weights remain unchanged throughout the pruning process.

\noindent \textbf{Evaluation protocol.}
We evaluate language modeling quality using perplexity on WikiText-2. 
For zero-shot downstream evaluation, we report accuracy on ARC-Challenge, ARC-Easy, BoolQ, HellaSwag, PIQA, and WinoGrande, and compute the average score across the six tasks. 
All reported zero-shot results are obtained without task-specific finetuning.

\section{Additional Sparsity Sensitivity Results}
\label{app:sparsity}
\begin{table}[h]
\centering
\scriptsize
\setlength{\tabcolsep}{3.2pt}
\renewcommand{\arraystretch}{0.95}
\caption{
Zero-shot downstream performance under different sparsity ratios.
We report results on six benchmarks and the average score.
}
\label{tab:sparsity_downstream}

\begin{tabular}{llccccccc}
\toprule
Model & Sparsity
& ARC-C & ARC-E & BoolQ & Hella. & PIQA & Wino. & Avg $\uparrow$ \\
\midrule

\multirow{2}{*}{LLaMA-2-7B}
& 30\% & 0.3481 & 0.6124 & 0.6606 & 0.6203 & 0.7258 & 0.6101 & 0.596 \\
& 40\% & 0.302 & 0.5109 & 0.5829 & 0.5394 & 0.6817 & 0.5738 & 0.532 \\

\midrule

\multirow{2}{*}{LLaMA-3-8B}
& 30\% & 0.3396 & 0.5934 & 0.6602 & 0.5913 & 0.7171 & 0.5809 & 0.580 \\
& 40\% & 0.2969 & 0.5004 & 0.6434 & 0.5092 & 0.6741 & 0.5509 & 0.529 \\

\bottomrule
\end{tabular}%
\vspace{-10pt}
\end{table}
Table~\ref{tab:sparsity_downstream} reports zero-shot results at 30\% and 40\% sparsity with $32\times32$ groups and WikiText-2 calibration.
Together with the 50\% results under the same setting (Tables~\ref{tab:ablation} and~\ref{tab:app_square_group_additional}), the average accuracy of both LLaMA-2-7B and LLaMA-3-8B decreases monotonically as sparsity increases from 30\% to 50\%.

\begin{table}[h]
\centering
\scriptsize
\setlength{\tabcolsep}{6pt}
\renewcommand{\arraystretch}{0.95}
\caption{
WikiText-2 perplexity under different sparsity ratios using the $32\times32$ group setting.
Lower values indicate better language modeling performance.
}
\label{tab:app_sparsity_ppl}

\begin{tabular}{lcc}
\toprule
Sparsity & LLaMA-2-7B $\downarrow$ & LLaMA-2-13B $\downarrow$ \\
\midrule
30\% & 7.32 & 5.98 \\
40\% & 8.25 & 6.79 \\
50\% & 9.37 & 7.79 \\
\bottomrule
\end{tabular}
\vspace{-8pt}
\end{table}
Table~\ref{tab:app_sparsity_ppl} reports WikiText-2 perplexity under the same setting.
For both LLaMA-2-7B and LLaMA-2-13B, perplexity increases monotonically as sparsity increases from 30\% to 50\%, consistent with the zero-shot results.

\begin{table}[tbh]
\centering
\scriptsize
\setlength{\tabcolsep}{3.2pt}
\renewcommand{\arraystretch}{0.92}
\caption{
Additional zero-shot downstream results under square group configurations.
}
\label{tab:app_square_group_additional}
\begin{tabular}{llccccccc}
\toprule
Group & Cal.
& ARC-C & ARC-E & BoolQ & Hella. & PIQA & Wino. & Avg $\uparrow$ \\
\midrule

\multicolumn{9}{l}{\textit{LLaMA-2-13B}} \\
\midrule
$32\times32$  & Wiki   & 0.3131 & 0.5640 & 0.6336 & 0.5621 & 0.6986 & 0.5691 & 0.557 \\
$32\times32$  & Alpaca & 0.3311 & 0.5800 & 0.6309 & 0.5486 & 0.7029 & 0.5801 & 0.562 \\
$64\times64$  & Wiki   & 0.3131 & 0.5505 & 0.5728 & 0.5600 & 0.6986 & 0.5762 & 0.545 \\
$64\times64$  & Alpaca & 0.3217 & 0.5703 & 0.6330 & 0.5505 & 0.6855 & 0.5833 & 0.557 \\
$128\times128$ & Wiki  & 0.3080 & 0.5122 & 0.5982 & 0.5654 & 0.6806 & 0.5706 & 0.539 \\
$128\times128$ & Alpaca & 0.3379 & 0.5319 & 0.6232 & 0.5592 & 0.6997 & 0.5643 & 0.553 \\

\midrule
\multicolumn{9}{l}{\textit{LLaMA-3-8B}} \\
\midrule
$32\times32$  & Wiki   & 0.2560 & 0.4415 & 0.6135 & 0.4143 & 0.6257 & 0.5351 & 0.481 \\
$32\times32$  & Alpaca & 0.2824 & 0.4954 & 0.5251 & 0.4283 & 0.6632 & 0.5335 & 0.488 \\
$64\times64$  & Wiki   & 0.2432 & 0.4108 & 0.6217 & 0.3933 & 0.6104 & 0.5588 & 0.473 \\
$64\times64$  & Alpaca & 0.2688 & 0.4461 & 0.5309 & 0.4027 & 0.6295 & 0.5454 & 0.471 \\
$128\times128$ & Wiki  & 0.2457 & 0.3864 & 0.6217 & 0.3775 & 0.6012 & 0.5422 & 0.462 \\
$128\times128$ & Alpaca & 0.2509 & 0.3956 & 0.6217 & 0.3713 & 0.6181 & 0.5272 & 0.464 \\

\midrule
\multicolumn{9}{l}{\textit{Qwen3-8B}} \\
\midrule
$32\times32$  & Wiki   & 0.3148 & 0.5227 & 0.6220 & 0.4214 & 0.6534 & 0.5288 & 0.511 \\
$32\times32$  & Alpaca & 0.2918 & 0.5105 & 0.6009 & 0.5480 & 0.6458 & 0.5114 & 0.518 \\
$64\times64$  & Wiki   & 0.2961 & 0.4726 & 0.6220 & 0.4167 & 0.6415 & 0.5296 & 0.496 \\
$64\times64$  & Alpaca & 0.3183 & 0.5303 & 0.5110 & 0.4141 & 0.6545 & 0.5391 & 0.495 \\
$128\times128$ & Wiki  & 0.2509 & 0.4251 & 0.6232 & 0.3870 & 0.6186 & 0.5328 & 0.473 \\
$128\times128$ & Alpaca & 0.2944 & 0.4874 & 0.6162 & 0.3983 & 0.6447 & 0.5280 & 0.495 \\

\midrule
\multicolumn{9}{l}{\textit{Qwen3-14B}} \\
\midrule
$32\times32$  & Wiki   & 0.3270 & 0.5644 & 0.5214 & 0.5038 & 0.6719 & 0.5706 & 0.527 \\
$32\times32$  & Alpaca & 0.3532 & 0.6090 & 0.6682 & 0.4947 & 0.6850 & 0.5588 & 0.561 \\
$64\times64$  & Wiki   & 0.3305 & 0.5535 & 0.4960 & 0.4896 & 0.6529 & 0.5549 & 0.513 \\
$64\times64$  & Alpaca & 0.3575 & 0.6157 & 0.5135 & 0.4809 & 0.6746 & 0.5659 & 0.535 \\
$128\times128$ & Wiki  & 0.3012 & 0.5021 & 0.6211 & 0.4542 & 0.6540 & 0.5406 & 0.512 \\
$128\times128$ & Alpaca & 0.3140 & 0.5400 & 0.6349 & 0.4647 & 0.6638 & 0.5359 & 0.526 \\

\bottomrule
\end{tabular}%
\vspace{-10pt}
\end{table}

\section{Square Group Granularity}
\label{app:square_group}

We provide additional zero-shot results for square group configurations in Table~\ref{tab:app_square_group_additional}.
These results complement the main ablation study by evaluating $32\times32$, $64\times64$, and $128\times128$ group shapes across multiple model families under both WikiText-2 and Alpaca calibration.

Overall, the results show that smaller square groups generally provide better pruning quality than coarser square groups.
For LLaMA-2-13B, the best square-group result is obtained by the $32\times32$ Alpaca setting, with an average score of 0.562.
For LLaMA-3-8B, $32\times32$ also gives the best square-group performance, with average scores of 0.481 using WikiText-2 calibration and 0.488 using Alpaca calibration.
A similar trend appears on Qwen3-8B, where $32\times32$ Alpaca achieves the highest average score among the square-group settings.
For Qwen3-14B, $32\times32$ Alpaca again performs best, reaching an average score of 0.561.

Larger square groups reduce the number of selection decisions: each $128\times128$ group covers 16 times as many weights as a $32\times32$ group.
Consistent with this, $128\times128$ underperforms $32\times32$ in every setting of Table~\ref{tab:app_square_group_additional}, and in Table~\ref{tab:ablation} the $1\times128$ and $1\times256$ configurations obtain higher averages than $32\times32$ and larger square groups.

\section{Additional Ablation and Efficiency Analysis}
\label{app:additional_speedup}

\begin{figure}[thb]
    \centering

    \begin{subfigure}[t]{0.32\linewidth}
        \centering
        \includegraphics[width=\linewidth]
        {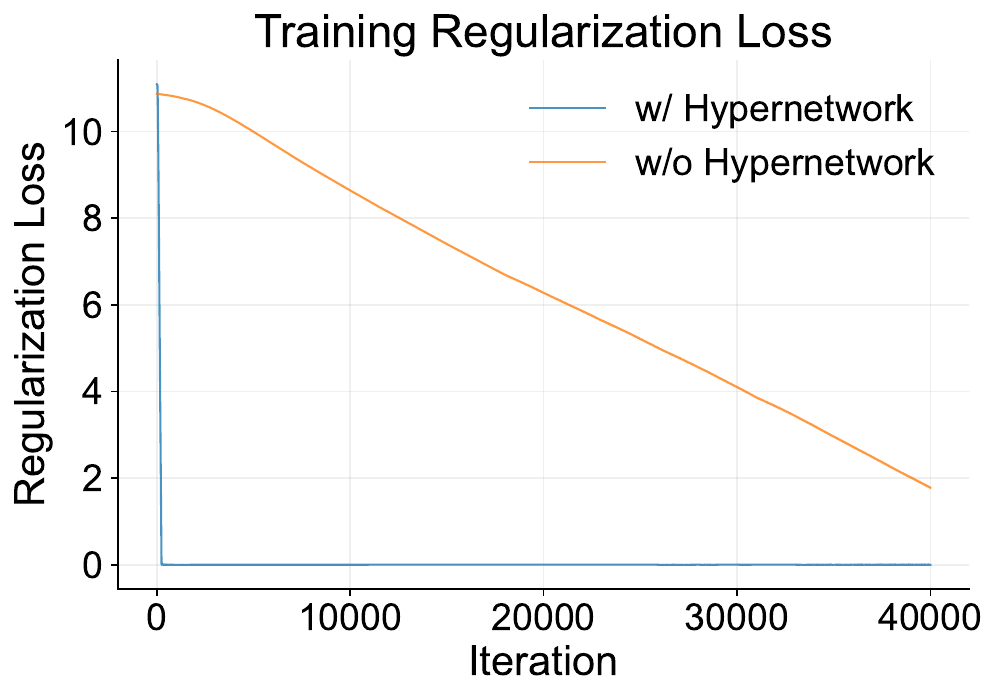}
        \caption{Regularization loss.}
        \label{fig:app_reg_loss}
    \end{subfigure}
    \hfill
    \begin{subfigure}[t]{0.32\linewidth}
        \centering
        \includegraphics[width=\linewidth]
        {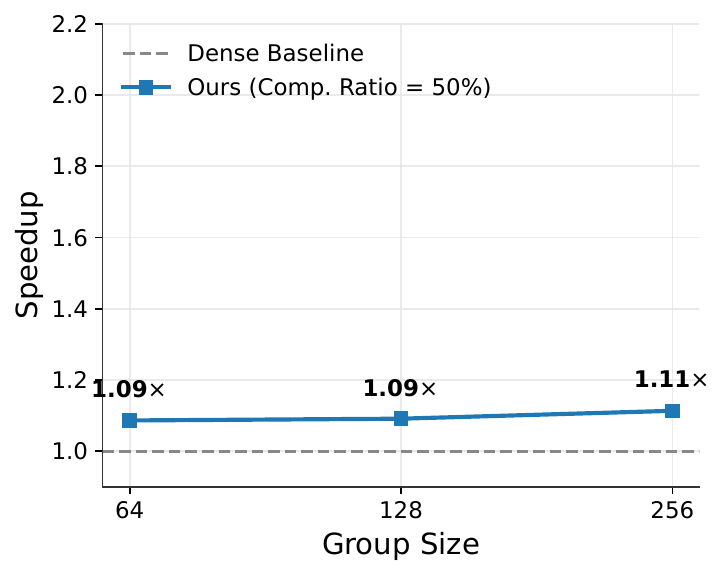}
        \caption{Speedup vs. group size.}
        \label{fig:app_speedup_group}
    \end{subfigure}
    \hfill
    \begin{subfigure}[t]{0.32\linewidth}
        \centering
        \includegraphics[width=\linewidth]
        {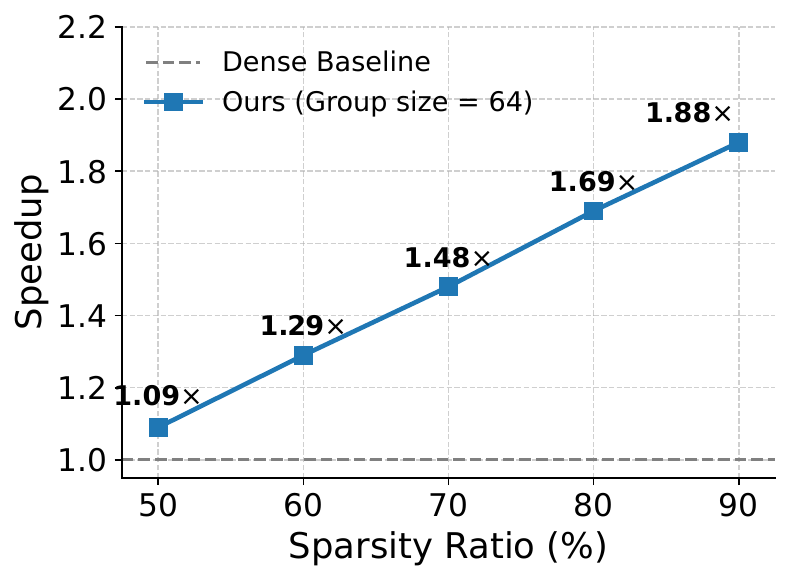}
        \caption{Speedup vs. Sparsity}
        \label{fig:app_efficiency}
    \end{subfigure}

    \caption{
    Additional analysis of GroupMask.
    (a) Training dynamics of the sparsity regularization loss with and
    without the hypernetwork.
    (b) Kernel speedup under different group sizes at 50\% sparsity.
    (c) Kernel speedup under group size 64 at different sparsity ratios.
    }
    \label{fig:additional_analysis}
\end{figure}
Figure~\ref{fig:app_reg_loss} shows that the sparsity regularization loss decreases faster with the hypernetwork than without it, complementing the task-loss comparison in Figure~\ref{fig:task_loss_comparison}.
Figure~\ref{fig:app_speedup_group} shows that, at 50\% sparsity, kernel speedup varies only slightly across the evaluated group sizes, and Figure~\ref{fig:app_efficiency} shows that speedup increases with sparsity.

\section{Baseline and Evaluation Details}

For baselines whose official results are available under the same setting, we report the published numbers.
For models or settings that are not reported by prior work and are not directly applicable, we leave the corresponding entries blank.
The symbol * indicates that the result is taken from the corresponding paper or follows its reported setting.
All zero-shot evaluations are conducted with lm-evaluation-harness using the official task metrics.

\section{Mask Export and Preserved Rate Computation}
\label{app:mask_export}

After mask learning, the continuous mask scores are converted into binary group selectors using the same thresholding rule as in evaluation.
The exported binary masks are then applied to the original pretrained weights to obtain the final sparse model.
For each projection matrix, the preserved rate is computed as the ratio between the number of active weights and the total number of weights after applying the binary masks.
For QKV projections, we report the average preserved rate over the query, key, and value projection matrices.
\end{document}